\documentclass[conference]{IEEEtran}
\IEEEoverridecommandlockouts
\usepackage{amsmath,amssymb,amsfonts}
\usepackage{algorithmic}
\usepackage{graphicx}
\usepackage{textcomp}

\usepackage[center]{caption}
\usepackage{graphicx}
\usepackage{xcolor}
\usepackage[
   sorting=nyt, % trier par nom, année, titre
   citestyle=numeric
   maxcitenames=2,
   maxbibnames=20,
   natbib=true,
   backend=bibtex, % compilateur par défaut pour biblatex
   style=numeric,
   sorting=none,
   isbn=false,
   url=false,
   doi=false,
   eprint=false,
   giveninits=true, % tous les prénoms sont des initiales
]{biblatex}
\usepackage{hyperref}
\bibliography{BiblEtat24} 

\def\BibTeX{{\rm B\kern-.05em{\sc i\kern-.025em b}\kern-.08em
    T\kern-.1667em\lower.7ex\hbox{E}\kern-.125emX}}
\begin{document}

\title{FABLE : Fabric Anomaly Detection Automation Process\\

}

\author{\IEEEauthorblockN{Simon Thomine}
\IEEEauthorblockA{
\textit{University of technology of Troyes}\\
\textit{AQUILAE} \\
Troyes, France \\
simon.thomine@utt.fr}
\and
\IEEEauthorblockN{Hichem Snoussi}
\IEEEauthorblockA{
\textit{University of technology of Troyes}\\
Troyes, France \\
hichem.snoussi@utt.fr}
\and
\IEEEauthorblockN{Mahmoud Soua}
\IEEEauthorblockA{\textit{AQUILAE} \\
Troyes, France \\
m.soua@aquilae.tech}
}

\maketitle

\begin{abstract}
Unsupervised anomaly in industry has been a concerning topic and a stepping stone for high performance industrial automation process. The vast majority of industry-oriented methods focus on learning from good samples to detect anomaly notwithstanding some specific industrial scenario requiring even less specific training and therefore a generalization for anomaly detection. The obvious use case is the fabric anomaly detection, where we have to deal with a really wide range of colors and types of textile and a stoppage of the production line for training could not be considered. In this paper, we propose an automation process for industrial fabric texture defect detection with a specificity-learning process during the domain-generalized anomaly detection. Combining the ability to generalize and the learning process offer a fast and precise anomaly detection and segmentation. The main contributions of this paper are the following: A domain-generalization texture anomaly detection method achieving the state-of-the-art performances, a fast specific training on good samples extracted by the proposed method, a self-evaluation method based on custom defect creation and an automatic detection of already seen fabric to prevent re-training.
\end{abstract}

\begin{IEEEkeywords}
Domain-Generalization, unsupervised, anomaly, unseen, knowledge distillation, student-teacher, memory banks, fabric, automation.
\end{IEEEkeywords}

\section{Introduction}
Unsupervised anomaly detection in industry is a vast topic, since there are a lot of
possible applications. In this paper, we focus on fabric anomaly, which is
a concerning topic for industry. The specificity of fabric is the pattern
in the structure and if we manage to understand that pattern we can extract
anomalies. Several methods have been introduced for industry anomaly detection using MVTEC AD \cite{bergmann_mvtec_2019} the dataset that gathers textures (carpet, leather,
grid, wood, and tile) and objects (bottle, cable, capsule, hazelnut, metal nut,
pill, screw, toothbrush, transistor and zipper). These methods could achieve
high performance. However, they rely on object/texture specific unsupervised learning without generalization capacity. Recently, knowledge-distillation based methods have been introduced
for the unsupervised anomaly detection task \cite{wang_student-teacher_2021}. It consists of a student-teacher model 
focusing on the bottom layers of the network as they represent the edges, color and shapes information.
We used the same approach to design a domain-generalized texture anomaly detection method with the ability to detect defects on unseen textures and to select good samples for a texture-specific unsupervised anomaly detection model. In fabric industry, many types and colors of fabric are analyzed, and it would be impossible to rely on a specific training on good samples for each type of fabric without slowing the industrial process. \\
Therefore, we propose a complete data processing chain for a robust, fast and adaptive texture specific anomaly detection and localization. Our method is based on four main modules: a domain-generalized texture anomaly detector, a fast texture specific training/inference, an auto-evaluation process of our specific model and an automatic already-seen fabric detection to avoid retraining an existing model. \\
The paper is organized as follows. In section II, we review the related work
especially on MVTEC dataset and present the different approaches proposed
in literature for domain-generalized and classic unsupervised anomaly detection. 
In section III, we present an enhanced domain-generalized texture defect detection method. In section IV, we present the specific learning method, the auto-evaluation process and the already seen texture recognition. Section V is dedicated to the analysis of the results. Section VI concludes the paper. 

\section{Related works}
As our proposed methods address two specific tasks, we first present the state of the art on domain-generalized texture anomaly detection and then the state of the art on unsupervised defect detection of known objects.
\subsection{Domain-generalized texture anomaly detection}
Domain-generalized anomaly detection is an important topic for optimal industrial process, since in specific industrial fields, the type of textures often changes. The most obvious example is certainly fabric anomaly detection where fabric can have different colors (red, blue, striped) and types (cotton, polyester, silk, etc). The main objective is to detect defects on any type of fabric without resorting to a time-consuming training. The feature extraction from a pretrained classifier offers the most promising results with different types of networks such as an episodic training \cite{li_episodic_2019}, the use of extrinsic and intrinsic aspects \cite{wang_learning_2020} and multiscale feature extractor with co-attention modules \cite{chen_domain-generalized_2022}. 

\subsection{Unsupervised anomaly detection on known objects}
More commonly, unsupervised anomaly detection deals with the problem of detecting defects on an object or texture based on only good samples. In industry or security scenarios, we often have a low rate of defects with a vast number of different defect types which would lead to a time-consuming annotation and a possibly non-pertinent classification if all the anomaly types are not considered\cite{han_adbench_2022}. To tackle this question, several methods emerged proposing different types of algorithms such as autoencoders  \cite{mei_automatic_2018} and  variational autoencoder variants \cite{nguyen_gee_2019} \cite{zavrtanik_draem_2021}. Another common way of detecting anomalies is Generative Adversarial Networks (GAN) introduced by \cite{goodfellow_generative_2014} adapted to unsupervised anomaly detection such as Ano-GAN \cite{schlegl_f-anogan_2019}, G2D \cite{pourreza_g2d_2021} and OCR-GAN \cite{liang_omni-frequency_2022}. More recently, approaches using a pretrained classifier has been at the heart of the research in industrial anomaly detection and offers outstanding performance. There are three main feature extraction-based approaches: normalizing flow, knowledge distillation and memory banks. The normalizing flow approach consists of a flow training based on relevant features of good samples from a pretrained network such as AlexNet \cite{he_deep_2015}, Resnet \cite{krizhevsky_imagenet_2017} or efficient-net \cite{tan_efficientnet_2020} trained on imageNet. Different strategies were used to enhance performance, such as a 2D flow \cite{yu_fastflow_2021} or a cross-scale flow \cite{rudolph_fully_2021}. Another interesting approach is the use of a memory bank to extract relevant information from different good samples and to use this memory bank to compare and detect if there is an anomaly \cite{roth_towards_2021}. Finally, the concept of knowledge distillation was adapted for unsupervised anomaly detection and localization \cite{wang_student-teacher_2021}. The idea is to train a student network based on the output features of a teacher (already pretrained for a classification purpose) and on good samples. The student will be able to reproduce teacher features on a good sample, but will not be as precise for a defective sample. Several methods used this principle with different strategies such as a multi-layer feature selection \cite{wang_student-teacher_2021}, an asymmetric student teacher \cite{rudolph_asymmetric_2022}, a coupled-hypersphere-based feature adaptation \cite{lee_cfa_2022} and a mixed-teacher approach \cite{thomine_mixed_2023}. 

\section{Knowledge distillation generalization}
The proposed model is based on the knowledge distillation framework, where a pretrained network is used as a teacher and a student network is trained to reproduce the teacher output on good samples.
% The idea of unsupervised anomaly detection knowledge distillation is to use a pretrained network as a teacher and to train a student to reproduce the teacher's output on good samples. 
The student network is then expected to not be able to reproduce teacher features on defective samples, a property which is used to detect abnormal samples. \\ 
For domain generalization, we propose to train the student  on different types of textures and using many teachers to guarantee generalization. In order to achieve this objective, we first constitute a new dataset based on fabric datasets \cite{leibe_fine-grained_2016} which regroups different categories of textures with different quality and homogeneity.  
\begin{figure}[h]
\centerline{\includegraphics[scale=0.2]{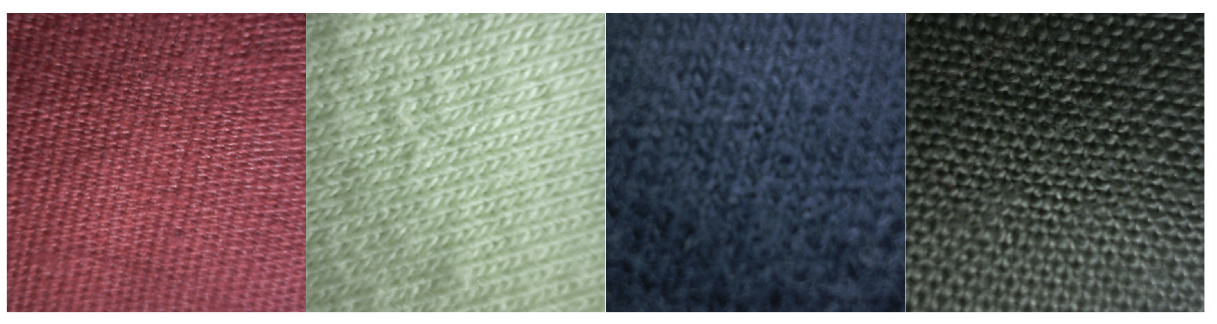}}
\renewcommand{\arraystretch}{1}
\caption{ 
Samples employed for the custom fabric dataset (extracted from the fabrics dataset \cite{leibe_fine-grained_2016})  }
\label{Fig.1}
\end{figure}

Then, to tackle the problem of texture domain generalization, we used a specific student teacher architecture with different branches based on the paradigm that each pre-trained classifier have a different bias towards classification. \\
In terms of layer selection, the deeper a layer, the more the information relates to the context and conversely, the shallower a layer, the more information it contains on contours, edges, and colors. Based on different layer configurations, we show that for the purpose of texture domain generalization, mid-level features would be the best choice to combine texture specific information such as contours and edges and a general vision of what a texture is. \\
% In the literature, the most commonly used pretrained classifier for anomaly detection purposes is Resnet followed by efficientNet both trained with imageNet. 
At least two classifiers are needed to attenuate each bias. We have used Resnet18 and EfficientNet-b0 for computation time speed and meaningful features. \\
To fully exploit each classifier information, we used a parallel architecture which can be seen as a multiple teachers/multiple students architecture where the training happen independently for each classifier, only the anomaly score is calculated with the two networks outputs.
Our framework is an adaptation of MixedTeacher \cite{thomine_mixed_2023} with a different layer selection strategy. The first Resnet layer is not used as its output features are too specific to training dataset textures. We used the features of the three first residual blocks of Resnet18 and the last 2 convolutional blocks of efficientNet-b0. As in \cite{thomine_mixed_2023}, we used a reduced version of the Resnet18 model with a reduction of the block size and a reduction of the dimension of each layer with an adaptive average pooling, while we keep the same architecture for the EfficientNet part.\\
 
\noindent Given a training dataset of images without anomaly ${D=[I_1,I_2,...,I_n]}$, our goal is to extract the information of $L$ mid-level layers. For an image ${I_k} \in R^{w*h*c}$ where $w$ is the width, $h$ the height, and $c$ the number of channel, the teacher outputs features $F_t^l(I_k) \in R^{w_l*h_l*c_l}$ and $F_s^l(I_k) \in R^{w_l/2*h_l/2*c_l/2}$ with $l>1 $ and $F_s^l(I_k) \in R^{w_l*h_l*c_l}$ if $l=1$. 
The loss is obtained by applying the $l2$ distance of normalized feature vectors for each pixel of the feature map and summing them.
% For the loss function, we took the $l2$ distance of normalized feature vectors for each pixel of the feature map and sum them all to obtain our global loss. 
For the Resnet student part, we used an adaptive average pooling layer on teacher features. The used layers are $l=\left\{ 1,2,3\right\}$ for the Resnet part and $l=\left\{5,6\right\}$ for the EfficientNet part.

\noindent Pixel loss  for the resnet part is defined in the following Eq.\ref{eq.1}: 

\small
\begin{equation}
loss^{l}(I_k)_{ij}=\frac{1}{2}\lVert norm(AAP(F_{Resnet18}^{l}(I_k))_{ij})-norm(F_s^l(I_k)_{ij}) \rVert 
\label{eq.1}
\end{equation}
\normalsize
\noindent where AAP refers to Adaptive Average Pooling. For the EfficientNet part, pixel loss  is defined in the following Eq.\ref{eq.2}: 
 
\begin{equation}
loss^{l}(I_k)_{ij}=\frac{1}{2}\lVert norm(F_{EffNetb0}^{l}(I_k)_{ij})-norm(F_s^l(I_k)_{ij}) \rVert 
\label{eq.2}
\end{equation}
\noindent For the layer $l$, the loss is defined as: 

\begin{equation}
loss^{l}(I_k)=\frac{1}{w_lh_l}  \sum_{i=1}^{w_l} \sum_{j=1}^{h_l} loss^l(I_k)_{ij}  
\label{eq.3}
\end{equation}

\noindent  and finally, for the total loss is written as: 

\begin{equation}
loss(I_k)= \sum^{l} loss^{l}(I_k) 
\label{eq.4}
\end{equation}

\begin{figure*}[h]
\centerline{\includegraphics[scale=0.22]{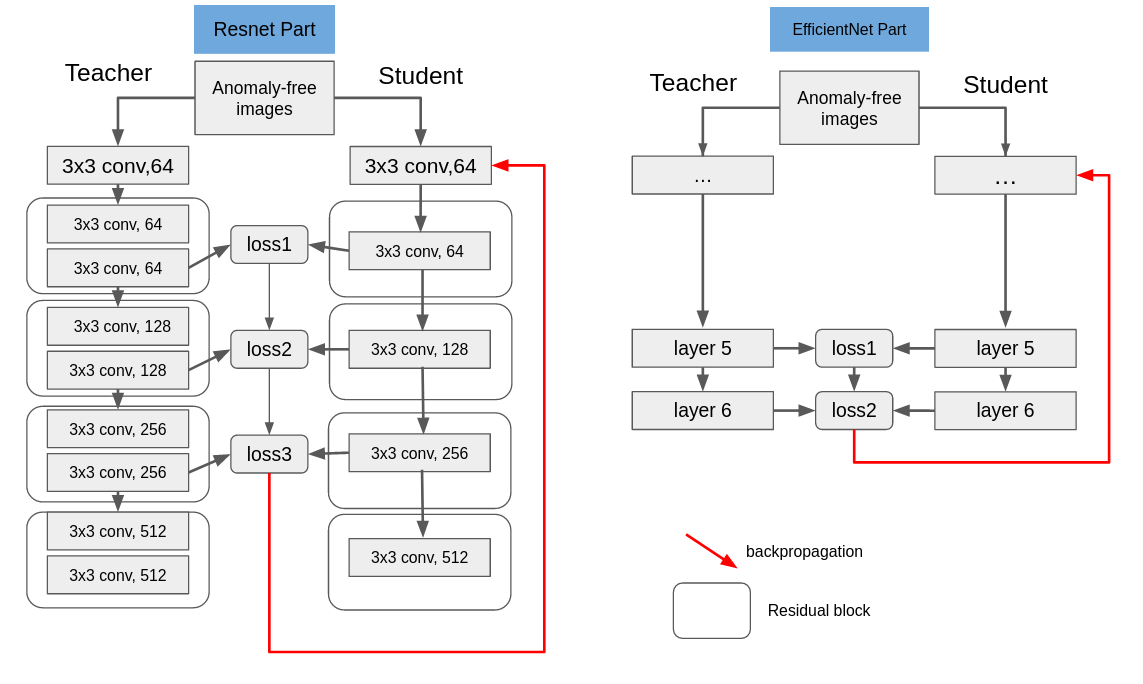}}
\renewcommand{\arraystretch}{1}
\caption{ 
Architecture of Resnet student teacher (left) and EfficientNet student teacher (right)}
\label{Fig.3}
\end{figure*}

\section{Auto-learning process for industrial deployment}
The previous part was presented in the context of industrial efficiency, where it was not allowed to retrain for every new type/color of texture/fabric. The objective of this section is to propose a general classifier for handling the anomaly detection role while we gather enough images and train a specific model for increased efficiency. \\
This section is divided in 3 parts: training and self-evaluation, recognition of an already trained type of fabric, and a typical industrial use-case in fabric industry.

\subsection{Training and self-evaluation}
Given the deployment constraints, we considered different criteria for the choice of the student-teacher network architecture: (i) the inference and training time, (ii) the performance and (iii) the robustness to defective samples in the training set. We also considered the possibility of running the process on several asynchronous defect detectors. The model Reduced Student proposed in \cite{thomine_mixed_2023} is a good candidate.  Thanks to its reduced architecture, we can train a specific model in an acceptable time. To minimize the number of potential defective samples in the training, we gathered the samples with acceptable anomaly score from the domain-generalized model, i.e samples classified as good samples. Based on a test-error approach, we determined the optimal number of epochs (during specific training) where the specific model becomes better than the domain-generalized one so that we can start using the best model even if the complete training is not finished. \\ 
The self-evaluation part is based on two types of data: (i) the first type is defective samples detected by the domain-generalized anomaly detector and (ii) the second type is generated data with a procedure inspired by DRAEM \cite{zavrtanik_draem_2021}: Perin noise and the texture database dtd \cite{cimpoi_describing_2014}. We used the same approach to generate non-absurd defects to self-evaluate our model. 
\begin{figure}[h]
\centerline{\includegraphics[scale=0.15]{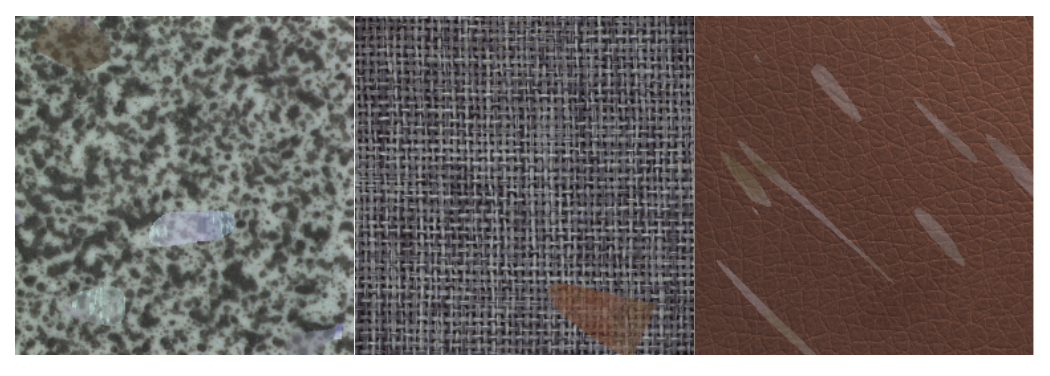}}
\renewcommand{\arraystretch}{1}
\caption{ 
Custom defective samples generated with Perin noise }
\label{Fig.2}
\end{figure}

\subsection{Already seen fabric recognition}
To guarantee an automated anomaly detector without the help of an operator for selecting an already-trained model, we propose an algorithm to precisely recognize a fabric type already considered previously. For each trained model, we save $x$ extracted features from the specific model on good samples reduced using coreset subsampling introduced in PatchCore \cite{roth_towards_2021} to guarantee fast computation. Each specific model is also saved in a model bank of $N$ models and linked to its features in a feature bank.
When we have to decide if the fabric was already seen, we calculate the sample/model proximity by extracting features from all trained specific models from the model bank, applying the coreset subsampling and comparing these features to the $x$ features from the feature bank of each specific model with cosine similarity distance as described in equation \ref{eq.6}. 
We then compute the intra-class proximity by calculating the cosine similarity between the $x$ features of the same model as reported in equation \ref{eq.7}.
The proximity score is defined as the absolute value of the difference between the sample/model proximity and the intra-class proximity.
We finally make the decision by comparing the maximum proximity score with a $similarityThreshold$. The threshold is chosen based on what is known about the similarity between the fabric. 
% We then compute the intra-class proximity by taking the mean of distances to all samples and the mean of each sample compared with the other $x-1$ extracted features to decide whether the new fabric is already-seen.
\begin{figure*}[h]
\centerline{\includegraphics[scale=0.25]{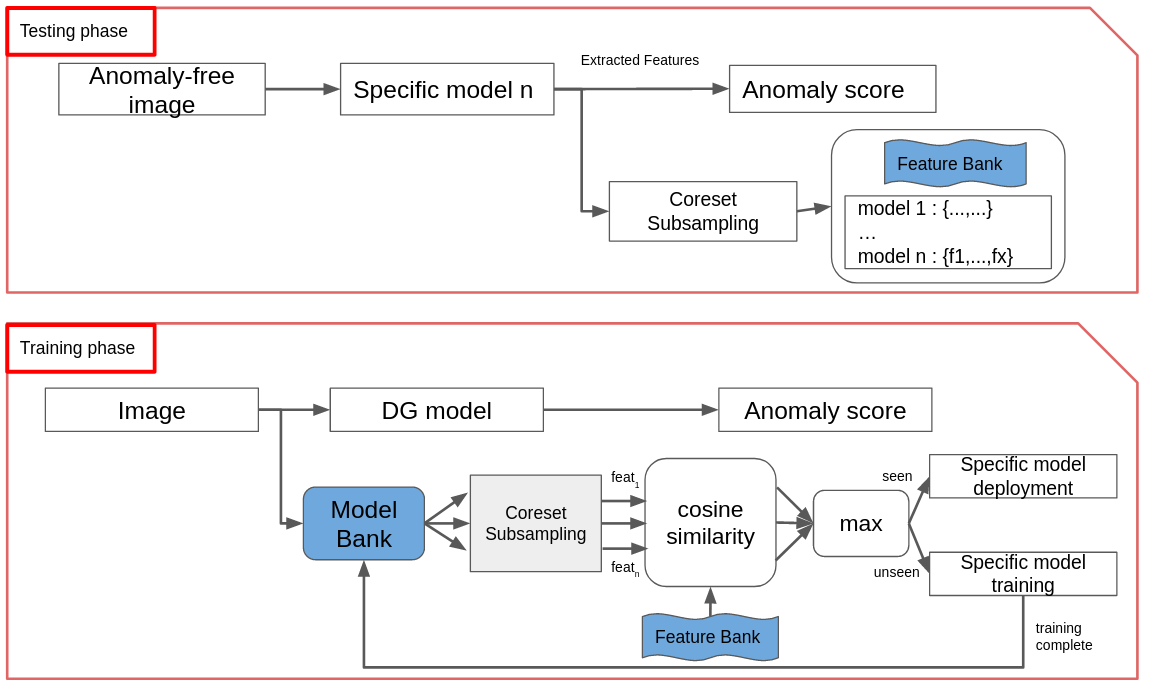}}
\renewcommand{\arraystretch}{1}
\caption{ 
Fabric Recognition process : The extraction of anomaly-free images features in the testing phase stops when the $x$ number of elements is reached, the model bank is the bank of all previously trained specific models}
\label{Fig.3}
\end{figure*}

Even though it may seem laborious if the model bank is consistent, it is still real-time deployable thanks to the inference speed of the reduced student architecture proposed in the previous subsection and the coreset subsampling, as we show in the experiment part. This is by far the most accurate method for comparing a new piece of fabric with a previously seen one and, we believe, it is still usable even in a specific case of thousands of specific models.
\\The cosine similarity formula is: 
\begin{equation}
sim(feat_A,feat_B)=\frac{feat_A.feat_B}{\left\|feat_A\right\|\left\|feat_B\right\|}
\label{eq.5} 
\end{equation}
with $feat_A$ and $feat_B$ the extracted features. The sample/model proximity is defined as: 
\begin{equation}
prox_{sm}(S,Model)=\frac{1}{x}  \sum_{i=1}^{x} sim(feat_S,feat_i)  
\label{eq.6} 
\end{equation}
The intra-class proximity is defined as:
\begin{equation}
prox_{ic}(Model)=\frac{1}{x(x-1)}  \sum_{i=1}^{x} \sum_{j=1,i\ne j}^{x} sim(feat_i,feat_j)  
\label{eq.7} 
\end{equation}
The proximity score is :
\small
\begin{equation}
proxScore(S,Model)= abs(prox_{sm}(S,Model) - prox_{ic}(Model))
\label{eq.8} 
\end{equation}
\normalsize

\noindent And the already-seen decision is described as  :
\begin{equation}
%Max_{i \in N}(proxScore(S,Model_i))> similarityThreshold
\underset{i \in N}{\text{max}}(proxScore(S,Model_i))> similarityThreshold
\label{eq.9} 
\end{equation}

\subsection{Typical use-case : fabric industry}
To demonstrate the effectiveness of our method, we describe a typical real defect analysis use case. In a vast majority of mid-range clothing industry, the fabric is analyzed several times during the whole fabrication process by operators that scroll the fabric and look for defects. This is a laborious job and often distraction occurs resulting in a globally low detection percentage of defects, not to mention the difficulty for the eyes to look at certain fabric categories such as striped fabrics. Our automated process aims at assisting the operator for the classification task and to speed up the scrolling of the fabric. The operator is still needed since he has to install the fabric roll on the machine and to verify the defect classification done by the domain generalized model since the accuracy is still low for a full automation process. \\
For every fabric roll, the process start with an identification of the fabric to control. Two different cases may happen: \\
- If this type of fabric has never been seen, a specific training is started while still doing the anomaly detection with the domain-generalized model, we may have to slow the scrolling of the fabric during the training depending on the computational power. When the trained model becomes better than the domain-generalized model, we used it instead, even if the training is not completely finished. When the training is finished, we used the completely trained model for anomaly detection while keeping some features of defective samples for the recognition part. \\
- If this type of fabric is already-seen, the specific trained model is used for anomaly detection. \\
The process is fully automated and does not require any help from the operator except for the activation or deactivation, which could be done also by connecting the visiting machine with the central unit to send an activation signal.

\section{Experiments}
This section is divided into 3 parts: the analysis of the domain-generalization model compared with state of the art for different training configurations, the analysis of the training speed and inference speed of our model and finally the estimation of the number of required epochs on a specific training to outperform the domain-generalization algorithm.\\

\subsection{State-of-the-art comparison}
For the evaluation of our model, we used two different databases for training. For the “MVTEC” one, we trained the DG model on all good samples of MVTEC AD textures except the one we are testing on to reproduce the evaluation protocol of the other state-of-the-art papers. The “cotton” one is trained on the custom fabric dataset presented in section III and was created for fabric anomaly which explain the SOTA performances on carpet and leather. The results are presented in table \ref{table:Table1}.\\
For the training, we used stochastic gradient descent with a learning rate of 0.4 for 200 epochs with a batch size of 16. Both networks are pretrained on ImageNet.
We resized all the images to 256x256, keeping 80\% for training and 20\% for validation. We kept the checkpoint with the lowest validation loss.
\begin{table}[h]
	\centering
	\caption{\textbf{AUC comparison between our method and existing ones on MVTEC AD}}
	\footnotesize
	\setlength{\tabcolsep}{11pt}
    \scalebox{0.65}{
	\begin{tabular}{|c|ccc|c|c|}
			
		\hline
		{\centering \textbf{textures}}  &
        {\textbf{Epi-FRC+\cite{li_episodic_2019}}}  &
        {\textbf{EISNet+\cite{wang_learning_2020}}}  &
        {\textbf{DGTSAD\cite{chen_domain-generalized_2022}}}  &
		{\textbf{Ours(MVTEC)}}  &
        {\textbf{Ours(Coton)}} \\
		\hline
		
		{carpet} &		
         {\centering 0.916} &
         {\centering 0.982} &
         {\centering 0.943} &
         {\centering 0.985} &
		{\centering \textbf{0.996}} \\
			
		{leather} &		
        {\centering 1.000} &
        {\centering 1.000} &
        {\centering \textbf{1.000}} &
        {\centering 0.991} &
		{\centering 0.996} 	\\
		
		{wood} &
        {\centering 0.941} &
        {\centering 0.979} &
        {\centering 0.962} &
        {\centering \textbf{0.999}} &
		{\centering 0.948} \\
		
		{tile} &
        {\centering 0.951} &
        {\centering 0.851} &
        {\centering \textbf{0.994}} &
        {\centering 0.965} &
		{\centering 0.964} \\
		
 		{grid} &	
        {\centering 0.725} &
        {\centering 0.728} &
        {\centering 0.730} &
        {\centering 0.937} &
 		{\centering \textbf{0.944}} \\ 		
		
        \hline
		{mean} &	
        {\centering 0.907} &
        {\centering 0.909} &
        {\centering 0.918} &
        {\centering \textbf{0.975}} &
		{\centering 0.969}\\	
        \hline
	\end{tabular} \label{table:Table1}
 }
\end{table}

\begin{figure}[h]
\centerline{\includegraphics[scale=0.15]{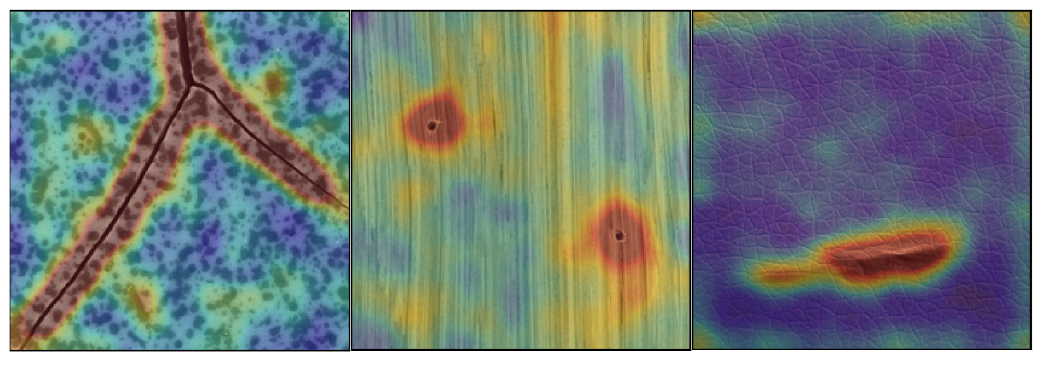}}
\renewcommand{\arraystretch}{1}
\caption{ 
Outputs of detected anomaly with our domain-generalization model }
\label{Fig.4}
\end{figure}

As seen in table \ref{table:Table1}, the 2 types of training offers approximately the same mean AUC but the dataset “cotton” only contains one hundred images and is supposed to be effective for fabric defects detection whereas it shows the best results on both carpet, leather and grid which contains patterns \cite{ngan_automated_2011} and are the most fabric-like textures of the dataset MVTEC AD.  \\
Comparing to the state-of-the-art methods, our approach obtains 0.057 AUC more than the previous best model, which is an excellent improvement and is in the way for closing the gap between domain generalization and classical anomaly detection for textures.

\subsection{Inference speed}
For the inference speed, all the following tests were done with a RTX 2080Ti. The training parameters are the same as the previous part except for the number of epochs where
we limited it to 100. To perform these experiments, we used an optimal algorithm for fast processing in batch of 8 which outperforms the classic anomalib \cite{akcay_anomalib_2022} in terms of inference speed on knowledge distillation methods for anomaly detection (see table  \ref{table:Table2}). For the training time, with 100 epochs and 100 images, we report 4 minutes and 20 seconds.
\begin{table}[h]
	\centering
	\caption{\textbf{Inference speed}}
	\footnotesize
	\setlength{\tabcolsep}{2pt}
	\scalebox{1.2}{
	\begin{tabular}{|c|cc|}
			
		\hline
		{\centering \textbf{Category}} & 
		{\textbf{Ours(Domain-Generalized)}} &
		{\textbf{Ours(Specific)}} \\
% 		&  &  & \\		
		\hline
		
		{FPS} &
		{\centering 333}  &
		{\centering \textbf{1111}}\\
		{Latency (ms)} &
		{\centering 3}  &
		{\centering \textbf{0.9}}\\
		\hline
	   % & &  &  \\
 		
	\end{tabular} \label{table:Table2}
	}
\end{table}

\subsection{Necessary training epochs before model replacement}
The main purpose of the whole method is to detect defects as precisely as possible during the whole process. We need to estimate at which point the specific model surpasses
the DG model in terms of AUC. In order to find this number, we tested our specific model at different epochs, and we compared results to the DG model performance.
\begin{table}[h]
	\centering
	\caption{\textbf{Epoch performance}}
	\footnotesize
	\setlength{\tabcolsep}{11pt}
    \scalebox{0.75}{
	\begin{tabular}{|c|ccc|c|c|}
			
		\hline
		{\centering \textbf{textures}}  &
        \textbf{10 epochs}  &
        \textbf{20 epochs}  &
        \textbf{30 epochs}  &
		\textbf{100 epochs}  &
        \textbf{DG model} \\
		\hline
		
		{carpet} &		
         {\centering 0.915} &
         {\centering 0.920} &
         {\centering 0.963} &
         {\centering 1.0} &
		{\centering 0.996} \\
			
		{leather} &		
        {\centering 0.949} &
        {\centering 0.974} &
        {\centering 0.990} &
        {\centering 0.997} &
		{\centering 0.996} 	\\
		
		{wood} &
        {\centering 0.988} &
        {\centering 0.989} &
        {\centering 0.995} &
        {\centering 0.996} &
		{\centering 0.948} \\
		
		{tile} &
        {\centering 0.986} &
        {\centering 0.987} &
        {\centering 0.991} &
        {\centering 0.987} &
		{\centering 0.964} \\

        \hline
		{mean} &	
        {\centering 0.959} &
        {\centering 0.967} &
        {\centering 0.984} &
        {\centering 0.995} &
		{\centering 0.969}\\	
        \hline
		
	\end{tabular} \label{table:Table3}
 }
\end{table}

Based on the mean results, we could consider using the specific model after 30 epochs of training i.e one minute and 30 seconds, but considering the scores of carpet and leather which are the most fabric-like textures, it may be a counter performance to switch to the specific model before the end of the training if the DG model is close enough to real distribution. 

\section{Conclusion and future work}
We proposed an industry-ready automation process for industrial defect detection, especially deployable for fabric with fast inference results for both domain-generalized and specific model. The outstanding capability of our architecture to mutual aid between humans and artificial intelligence makes it  a great and reliable tool for visual inspection.
Nevertheless, several improvements could be considered. According to \cite{han_adbench_2022}, semi-supervised can easily surpass unsupervised anomaly detection even with a few annotated anomalies and this could be applied to our method by using the DG detected anomalies to train a semi supervised specific model to increase the detection performances. 
Some automation improvement could be considered in terms of potential use of the method, such as a direct software included in the fabric visiting machine to monitor speed, stop and dysfunction mode more precisely.

% Parler du semi-supervised learning
\small
\printbibliography

@article{goodfellow_generative_2014,
	title = {Generative Adversarial Networks},
	url = {http://arxiv.org/abs/1406.2661},
	journaltitle = {{arXiv}:1406.2661 [cs, stat]},
	author = {Goodfellow, Ian J. and Pouget-Abadie, Jean and Mirza, Mehdi and Xu, Bing and Warde-Farley, David and Ozair, Sherjil and Courville, Aaron and Bengio, Yoshua},
	urldate = {2021-06-15},
	date = {2014-06-10},
	eprinttype = {arxiv},
	eprint = {1406.2661},
}

@inproceedings{pourreza_g2d_2021,
	title = {G2D: Generate to Detect Anomaly},
	isbn = {978-1-66540-477-8},
	url = {https://ieeexplore.ieee.org/document/9423181/},
	doi = {10.1109/WACV48630.2021.00205},
	shorttitle = {G2D},
	eventtitle = {2021 {IEEE} Winter Conference on Applications of Computer Vision ({WACV})},
	pages = {2002--2011},
	booktitle = {2021 {IEEE} Winter Conference on Applications of Computer Vision ({WACV})},
	publisher = {{IEEE}},
	author = {Pourreza, Masoud and Mohammadi, Bahram and Khaki, Mostafa and Bouindour, Samir and Snoussi, Hichem and Sabokrou, Mohammad},
	urldate = {2021-06-15},
	date = {2021-01},
	note = {event-place: Waikoloa, {HI}, {USA}},
}

@article{nguyen_gee_2019,
	title = {{GEE}: A Gradient-based Explainable Variational Autoencoder for Network Anomaly Detection},
	url = {http://arxiv.org/abs/1903.06661},
	shorttitle = {{GEE}},
	journaltitle = {{arXiv}:1903.06661 [cs, stat]},
	author = {Nguyen, Quoc Phong and Lim, Kar Wai and Divakaran, Dinil Mon and Low, Kian Hsiang and Chan, Mun Choon},
	urldate = {2021-06-16},
	date = {2019-03-15},
	eprinttype = {arxiv},
	eprint = {1903.06661},
}

@article{mei_automatic_2018,
	title = {Automatic Fabric Defect Detection with a Multi-Scale Convolutional Denoising Autoencoder Network Model},
	volume = {18},
	url = {https://www.mdpi.com/1424-8220/18/4/1064},
	doi = {10.3390/s18041064},
	pages = {1064},
	number = {4},
	journaltitle = {Sensors},
	author = {Mei, Shuang and Wang, Yudan and Wen, Guojun},
	urldate = {2021-07-07},
	date = {2018-04},
}

@article{schlegl_f-anogan_2019,
	title = {f-{AnoGAN}: Fast unsupervised anomaly detection with generative adversarial networks},
	volume = {54},
	issn = {13618415},
	url = {https://linkinghub.elsevier.com/retrieve/pii/S1361841518302640},
	doi = {10.1016/j.media.2019.01.010},
	shorttitle = {f-{AnoGAN}},
	pages = {30--44},
	journaltitle = {Medical Image Analysis},
	shortjournal = {Medical Image Analysis},
	author = {Schlegl, Thomas and Seeböck, Philipp and Waldstein, Sebastian M. and Langs, Georg and Schmidt-Erfurth, Ursula},
	urldate = {2021-07-09},
	date = {2019-05},
}

@article{roth_towards_2021,
	title = {Towards Total Recall in Industrial Anomaly Detection},
	url = {http://arxiv.org/abs/2106.08265},
	journaltitle = {{arXiv}:2106.08265 [cs]},
	author = {Roth, Karsten and Pemula, Latha and Zepeda, Joaquin and Schölkopf, Bernhard and Brox, Thomas and Gehler, Peter},
	urldate = {2021-10-14},
	date = {2021-06-15},
	eprinttype = {arxiv},
	eprint = {2106.08265},
}

@article{yu_fastflow_2021,
	title = {{FastFlow}: Unsupervised Anomaly Detection and Localization via 2D Normalizing Flows},
	url = {http://arxiv.org/abs/2111.07677},
	shorttitle = {{FastFlow}},
	journaltitle = {{arXiv}:2111.07677 [cs]},
	author = {Yu, Jiawei and Zheng, Ye and Wang, Xiang and Li, Wei and Wu, Yushuang and Zhao, Rui and Wu, Liwei},
	urldate = {2021-11-22},
	date = {2021-11-16},
	eprinttype = {arxiv},
	eprint = {2111.07677},
}

@article{rudolph_fully_2021,
	title = {Fully Convolutional Cross-Scale-Flows for Image-based Defect Detection},
	url = {http://arxiv.org/abs/2110.02855},
	journaltitle = {{arXiv}:2110.02855 [cs]},
	author = {Rudolph, Marco and Wehrbein, Tom and Rosenhahn, Bodo and Wandt, Bastian},
	urldate = {2021-11-29},
	date = {2021-10-06},
	eprinttype = {arxiv},
	eprint = {2110.02855},
}

@article{liang_omni-frequency_2022,
	title = {Omni-frequency Channel-selection Representations for Unsupervised Anomaly Detection},
	url = {http://arxiv.org/abs/2203.00259},
	journaltitle = {{arXiv}:2203.00259 [cs]},
	author = {Liang, Yufei and Zhang, Jiangning and Zhao, Shiwei and Wu, Runze and Liu, Yong and Pan, Shuwen},
	urldate = {2022-05-05},
	date = {2022-03-01},
	langid = {english},
	eprinttype = {arxiv},
	eprint = {2203.00259},
}

@article{wang_student-teacher_2021,
	title = {Student-Teacher Feature Pyramid Matching for Anomaly Detection},
	url = {http://arxiv.org/abs/2103.04257},
	journaltitle = {{arXiv}:2103.04257 [cs]},
	author = {Wang, Guodong and Han, Shumin and Ding, Errui and Huang, Di},
	urldate = {2022-05-05},
	date = {2021-10-28},
	langid = {english},
	eprinttype = {arxiv},
	eprint = {2103.04257},
}

@inproceedings{bergmann_mvtec_2019,
	location = {Long Beach, {CA}, {USA}},
	title = {{MVTec} {AD} — A Comprehensive Real-World Dataset for Unsupervised Anomaly Detection},
	isbn = {978-1-72813-293-8},
	url = {https://ieeexplore.ieee.org/document/8954181/},
	doi = {10.1109/CVPR.2019.00982},
	eventtitle = {2019 {IEEE}/{CVF} Conference on Computer Vision and Pattern Recognition ({CVPR})},
	pages = {9584--9592},
	booktitle = {2019 {IEEE}/{CVF} Conference on Computer Vision and Pattern Recognition ({CVPR})},
	publisher = {{IEEE}},
	author = {Bergmann, Paul and Fauser, Michael and Sattlegger, David and Steger, Carsten},
	urldate = {2022-05-11},
	date = {2019-06},
	langid = {english},
}

@incollection{leibe_fine-grained_2016,
	location = {Cham},
	title = {Fine-Grained Material Classification Using Micro-geometry and Reflectance},
	volume = {9909},
	isbn = {978-3-319-46453-4 978-3-319-46454-1},
	url = {http://link.springer.com/10.1007/978-3-319-46454-1_47},
	pages = {778--792},
	booktitle = {Computer Vision – {ECCV} 2016},
	publisher = {Springer International Publishing},
	author = {Kampouris, Christos and Zafeiriou, Stefanos and Ghosh, Abhijeet and Malassiotis, Sotiris},
	editor = {Leibe, Bastian and Matas, Jiri and Sebe, Nicu and Welling, Max},
	urldate = {2022-05-16},
	date = {2016},
	langid = {english},
	doi = {10.1007/978-3-319-46454-1_47},
	note = {Series Title: Lecture Notes in Computer Science},
}

@misc{lee_cfa_2022,
	title = {{CFA}: Coupled-hypersphere-based Feature Adaptation for Target-Oriented Anomaly Localization},
	url = {http://arxiv.org/abs/2206.04325},
	shorttitle = {{CFA}},
	number = {{arXiv}:2206.04325},
	publisher = {{arXiv}},
	author = {Lee, Sungwook and Lee, Seunghyun and Song, Byung Cheol},
	urldate = {2022-09-12},
	date = {2022-06-09},
	langid = {english},
	eprinttype = {arxiv},
	eprint = {2206.04325 [cs]},
}

@misc{he_deep_2015,
	title = {Deep Residual Learning for Image Recognition},
	url = {http://arxiv.org/abs/1512.03385},
	number = {{arXiv}:1512.03385},
	publisher = {{arXiv}},
	author = {He, Kaiming and Zhang, Xiangyu and Ren, Shaoqing and Sun, Jian},
	urldate = {2022-09-30},
	date = {2015-12-10},
	langid = {english},
	eprinttype = {arxiv},
	eprint = {1512.03385 [cs]},
}

@misc{tan_efficientnet_2020,
	title = {{EfficientNet}: Rethinking Model Scaling for Convolutional Neural Networks},
	url = {http://arxiv.org/abs/1905.11946},
	shorttitle = {{EfficientNet}},
	number = {{arXiv}:1905.11946},
	publisher = {{arXiv}},
	author = {Tan, Mingxing and Le, Quoc V.},
	urldate = {2022-09-30},
	date = {2020-09-11},
	langid = {english},
	eprinttype = {arxiv},
	eprint = {1905.11946 [cs, stat]},
}

@misc{rudolph_asymmetric_2022,
	title = {Asymmetric Student-Teacher Networks for Industrial Anomaly Detection},
	url = {http://arxiv.org/abs/2210.07829},
	number = {{arXiv}:2210.07829},
	publisher = {{arXiv}},
	author = {Rudolph, Marco and Wehrbein, Tom and Rosenhahn, Bodo and Wandt, Bastian},
	urldate = {2022-11-15},
	date = {2022-10-18},
	langid = {english},
	eprinttype = {arxiv},
	eprint = {2210.07829 [cs]},
}

@misc{zavrtanik_draem_2021,
	title = {{DRAEM} -- A discriminatively trained reconstruction embedding for surface anomaly detection},
	url = {http://arxiv.org/abs/2108.07610},
	number = {{arXiv}:2108.07610},
	publisher = {{arXiv}},
	author = {Zavrtanik, Vitjan and Kristan, Matej and Skočaj, Danijel},
	urldate = {2022-11-21},
	date = {2021-09-27},
	langid = {english},
	eprinttype = {arxiv},
	eprint = {2108.07610 [cs]},
}

@misc{wang_learning_2020,
	title = {Learning from Extrinsic and Intrinsic Supervisions for Domain Generalization},
	url = {http://arxiv.org/abs/2007.09316},
	number = {{arXiv}:2007.09316},
	publisher = {{arXiv}},
	author = {Wang, Shujun and Yu, Lequan and Li, Caizi and Fu, Chi-Wing and Heng, Pheng-Ann},
	urldate = {2022-11-24},
	date = {2020-07-17},
	langid = {english},
	eprinttype = {arxiv},
	eprint = {2007.09316 [cs]},
}

@inproceedings{li_episodic_2019,
	location = {Seoul, Korea (South)},
	title = {Episodic Training for Domain Generalization},
	isbn = {978-1-72814-803-8},
	url = {https://ieeexplore.ieee.org/document/9008109/},
	doi = {10.1109/ICCV.2019.00153},
	eventtitle = {2019 {IEEE}/{CVF} International Conference on Computer Vision ({ICCV})},
	pages = {1446--1455},
	booktitle = {2019 {IEEE}/{CVF} International Conference on Computer Vision ({ICCV})},
	publisher = {{IEEE}},
	author = {Li, Da and Zhang, Jianshu and Yang, Yongxin and Liu, Cong and Song, Yi-Zhe and Hospedales, Timothy},
	urldate = {2022-11-24},
	date = {2019-10},
	langid = {english},
}

@misc{chen_domain-generalized_2022,
	title = {Domain-Generalized Textured Surface Anomaly Detection},
	url = {http://arxiv.org/abs/2203.12304},
	number = {{arXiv}:2203.12304},
	publisher = {{arXiv}},
	author = {Chen, Shang-Fu and Liu, Yu-Min and Lin, Chia-Ching and Chen, Trista Pei-Chun and Wang, Yu-Chiang Frank},
	urldate = {2022-11-24},
	date = {2022-03-23},
	langid = {english},
	eprinttype = {arxiv},
	eprint = {2203.12304 [cs]},
}

@article{krizhevsky_imagenet_2017,
	title = {{ImageNet} classification with deep convolutional neural networks},
	volume = {60},
	issn = {0001-0782, 1557-7317},
	url = {https://dl.acm.org/doi/10.1145/3065386},
	doi = {10.1145/3065386},
	pages = {84--90},
	number = {6},
	journaltitle = {Communications of the {ACM}},
	shortjournal = {Commun. {ACM}},
	author = {Krizhevsky, Alex and Sutskever, Ilya and Hinton, Geoffrey E.},
	urldate = {2022-11-25},
	date = {2017-05-24},
	langid = {english},
}

@inproceedings{cimpoi_describing_2014,
	location = {Columbus, {OH}, {USA}},
	title = {Describing Textures in the Wild},
	isbn = {978-1-4799-5118-5},
	url = {https://ieeexplore.ieee.org/document/6909856},
	doi = {10.1109/CVPR.2014.461},
	eventtitle = {2014 {IEEE} Conference on Computer Vision and Pattern Recognition ({CVPR})},
	pages = {3606--3613},
	booktitle = {2014 {IEEE} Conference on Computer Vision and Pattern Recognition},
	publisher = {{IEEE}},
	author = {Cimpoi, Mircea and Maji, Subhransu and Kokkinos, Iasonas and Mohamed, Sammy and Vedaldi, Andrea},
	urldate = {2022-12-01},
	date = {2014-06},
	langid = {english},
}

@misc{han_adbench_2022,
	title = {{ADBench}: Anomaly Detection Benchmark},
	url = {http://arxiv.org/abs/2206.09426},
	shorttitle = {{ADBench}},
	number = {{arXiv}:2206.09426},
	publisher = {{arXiv}},
	author = {Han, Songqiao and Hu, Xiyang and Huang, Hailiang and Jiang, Mingqi and Zhao, Yue},
	urldate = {2022-12-01},
	date = {2022-09-16},
	langid = {english},
	eprinttype = {arxiv},
	eprint = {2206.09426 [cs]},
}

@misc{akcay_anomalib_2022,
	title = {Anomalib: A Deep Learning Library for Anomaly Detection},
	url = {http://arxiv.org/abs/2202.08341},
	shorttitle = {Anomalib},
	number = {{arXiv}:2202.08341},
	publisher = {{arXiv}},
	author = {Akcay, Samet and Ameln, Dick and Vaidya, Ashwin and Lakshmanan, Barath and Ahuja, Nilesh and Genc, Utku},
	urldate = {2022-12-02},
	date = {2022-02-16},
	langid = {english},
	eprinttype = {arxiv},
	eprint = {2202.08341 [cs]},
}

@inproceedings{thomine_mixed_2023,
	title = {MixedTeacher : Knowledge Distillation for fast inference textural
anomaly detection},
	shorttitle = {MixedTeacher},

	author = {Thomine, Simon and Snoussi, Hichem and Soua, Mahmoud},
	urldate = {2023-02-19},
	date = {2023-02-19},
	langid = {english},
    eventtitle = {VISAPP international conference on computer vision theory and applications}
}

@article{ngan_automated_2011,
	title = {Automated fabric defect detection—A review},
	volume = {29},
	issn = {0262-8856},
	url = {https://www.sciencedirect.com/science/article/pii/S0262885611000230},
	doi = {10.1016/j.imavis.2011.02.002},
	pages = {442--458},
	number = {7},
	journaltitle = {Image and Vision Computing},
	shortjournal = {Image and Vision Computing},
	author = {Ngan, Henry Y. T. and Pang, Grantham K. H. and Yung, Nelson H. C.},
	urldate = {2023-01-02},
	date = {2011-06-01},
	langid = {english},
}

\end{document}